\documentclass[11pt]{article}

 \usepackage[letterpaper, left=1in, right=1in, top=0.95in, bottom=0.95in]{geometry}
 \usepackage{tikz}

\usepackage[numbers,square]{natbib}

\usepackage[utf8]{inputenc} 
\usepackage[T1]{fontenc}    
\usepackage[colorlinks=true, linkcolor=black, citecolor=blue, urlcolor=black, filecolor=blue, pdfborder={0 0 0}]{hyperref}       
\usepackage{url}            
\usepackage{booktabs}       
\usepackage{amsfonts,amsmath,amsthm}       
\usepackage{nicefrac}       
\usepackage{microtype}      
\usepackage{xcolor}         
\usepackage[capitalize,noabbrev]{cleveref}
 \usepackage{thm-restate}
 \usepackage{subcaption}
 
\title{The Impossibility of Inverse Permutation Learning in Transformer Models}

\newtheorem{lemma}{Lemma}

\newtheorem{definition}{Definition}

\crefname{task}{Task}{Tasks}

\newcommand{\embed}{\mathtt{embed}}
\newcommand{\MASK}{\mathrm{MASK}}
\newcommand{\attn}{\mathrm{attn}}
\newcommand{\R}{\mathbb{R}}
\newcommand{\sqparen}[1]{\left[ {#1} \right]}
\newcommand{\cS}{\mathcal{S}}

\newcommand{\1}{\mathbf{1}}
\newcommand{\activations}[1]{\cS(\MASK({#1}))}

\newcommand{\defeq}{\overset{\mathrm{def.}}{=}}

\author{
  Rohan Alur$^*$\\
  Bridgewater AIA Labs \\
  Massachusetts Institute of Technology \\
  \texttt{ralur@mit.edu} \\
  \and
  Chris Hays$^*$ \\
  Massachusetts Institute of Technology \\
  \texttt{jhays@mit.edu} \\
  \and
  Manish Raghavan \\
   Massachusetts Institute of Technology \\
  \texttt{mragh@mit.edu} \\
  \and
  Devavrat Shah \\
   Massachusetts Institute of Technology \\
  \texttt{devavrat@mit.edu} \\
}

\title{The Impossibility of Inverse Permutation Learning in Transformer Models}

\date{}

\begin{document}

\maketitle
\def\thefootnote{*}\footnotetext{These authors contributed equally to this work}\def\thefootnote{\arabic{footnote}}

\begin{abstract}
   In this technical note, we study the problem of inverse permutation learning in decoder-only transformers. Given a permutation and a string to which that permutation has been applied, the model is tasked with producing the original (``canonical'') string. We argue that this task models a natural robustness property across a variety of reasoning tasks, including long-context retrieval, multiple choice QA and in-context learning. Our primary contribution is an impossibility result: we show that an arbitrary depth, decoder-only transformer cannot learn this task. This result concerns the expressive capacity of decoder-only transformer models and is agnostic to training dynamics or sample complexity. We give a pair of alternative constructions under which inverse permutation learning is feasible. The first of these highlights the fundamental role of the causal attention mask, and reveals a gap between the expressivity of encoder-decoder transformers and the more popular decoder-only architecture. The latter result is more surprising: we show that simply padding the input with ``scratch tokens" yields a construction under which inverse permutation learning is possible. We conjecture that this may suggest an alternative mechanism by which chain-of-thought prompting or, more generally, intermediate ``thinking'' tokens can enable reasoning in large language models, even when these tokens encode no meaningful semantic information (e.g., the results of intermediate computations).
   
\end{abstract}

\section{Introduction}
\label{sec:intro}
One of the most striking features of modern large language models (LLMs) is the emergence of general-purpose reasoning abilities at scale. Even relatively early LLMs were capable of in-context learning \cite{brown2020fewshot}, multiple choice question answering \cite{Hendrycks2020-uh}, long-context reasoning \cite{Liu2023-nb}, and deductive logical reasoning tasks \cite{Han2022-hm}. It was not obvious, a priori, that a single model could be trained to handle such a diverse range of tasks, since each might appear to demand distinct—and potentially incompatible—architectures, training recipes or inductive biases.

In this work, we focus on a particular inductive bias: permutation invariance. Modern LLMs typically encode positional information through positional encodings \cite{vaswani2017attention}. This is, of course, a desirable property for a language model, as the order of words is inextricably linked to their meaning. However, this sensitivity to ordering can present challenges in other contexts. For example, consider in-context learning: given $k$ labeled examples, a model is asked to predict the label for a final unlabeled input. It is typically desirable that these predictions are \emph{invariant} to ordering of the examples. In multiple choice question answering (QA), we'd similarly like invariance to the order of answer options; in long-context reasoning, we might want invariance to the order of in-context facts; in deductive reasoning tasks a model should be invariant to the ordering of logical predicates; in generative verification tasks, the verifier should be invariant to the ordering of candidate solutions.

Unfortunately, modern LLMs fail to satisfy any of these desiderata \cite{Pezeshkpour2023-ct, Chen2024-ox, Liu2023-ot, Shi2024-ar, Lu2021-lz, Li2023-kk}. This sensitivity to ordering manifests as seemingly arbitrary failures, undermining reliability and, ultimately, trust in these models. This problem is well-studied, with proposed solutions including altering the training loss to encourage permutation invariance \cite{karlik2020invariance}, optimizing input ordering \cite{guo-etal-2024-makes, Bhope2025OptiSeqOE, atul2025opti}, and modifications to the architecture \cite{Egressy2025-zv, Xu2023-zh, gardner2024tabula}. However, none of these solutions address the fundamental question: can a standard decoder-transformer \emph{learn} to guarantee permutation invariance?

\textbf{Inverse permutation learning.} We observe that \emph{inverse permutation learning} is a sufficient and natural manner in which to guarantee order invariance. For example, given a sequence of (key, value) pairs as in-context examples, it might first map these examples to a lexicographic ordering over keys, guaranteeing invariance to the order in which they are presented. We model this task as \emph{inverse permutation learning}: given a permuted sequence and a description of the permutation which was applied, the model should learn to output a ``canonical'' version of the sequence. 

\begin{figure}[h]
\centering
\begin{tikzpicture}[x=1cm,y=1cm]
  \node[font=\footnotesize] at (0,0.75) {\textbf{Permutation}};
  \node[font=\footnotesize] at (3,0.75) {\textbf{Permuted Sequence}};
  \node[font=\footnotesize] at (7,0.75) {\textbf{Canonical Ordering}};

  \node (perm1) at (0,0) {$[\,2\;0\;1\,]$};
  \node (inp1)  at (3,0) {$[\,\clubsuit\;\heartsuit\;\spadesuit\,]$};
  \node (out1)  at (7,0) {$[\,\heartsuit\;\spadesuit\;\clubsuit\,]$};

  \draw[->,thick] (inp1.east) -- (out1.west)
    node[midway,above,sloped,font=\footnotesize]{apply inverse};

  \node (perm2) at (0,-1) {$[\,1\;2\;0\,]$};
  \node (inp2)  at (3,-1) {$[\,\spadesuit\;\clubsuit\;\heartsuit\,]$};
  \node (out2)  at (7,-1) {$[\,\heartsuit\;\spadesuit\;\clubsuit\,]$};

  \draw[->,thick] (inp2.east) -- (out2.west)
    node[midway,above,sloped,font=\footnotesize]{apply inverse};
\end{tikzpicture}
\caption{Visualizing inverse permutation learning. The model is given a permutation and a permuted sequence as input, and is tasked with inverting the permutation to recover the canonical ordering.}
\end{figure}

This task is perhaps easier than our motivating examples, as the model is given both a description of the permutation to be ``undone'' as well as the permuted sequence. Nonetheless, we show that a simplified transformer model cannot learn to complete this task.

\textbf{Contributions.} In \cref{sec:impossibility}, we derive an impossibility result for inverse permutation learning with a decoder-only, attention-only transformer architecture. The intuition underpinning this result is straightforward: all permutations (except the trivial identity permutation) require shifting at least one element from a later position in the sequence to an earlier position in the sequence. However, the structure of the causal attention mask prevents the necessary information transfer from later tokens to earlier ones. 

We complement this impossibility result with a pair of existence proofs in \Cref{sec:possibility_nocausal} and \Cref{sec:possibility_copy}. In \cref{sec:possibility_nocausal}, we demonstrate that removing causal masking to allow for general two-way attention mechanisms yields a construction which solves the inverse permutation learning problem. This result suggests that alternative architectures, particularly encoder-decoder architectures, may not have the same limitations as the decoder-only architecture which underpins the most popular modern LLMs. We corroborate this result with an empirical analysis in simplified transformer model, which demonstrates that accuracy in the inverse permutation learning task improves from the level of random guessing to nearly perfect accuracy if the causal attention mask is removed.\footnote{Our code is available here: \href{https://github.com/johnchrishays/icl}{https://github.com/johnchrishays/icl}.} We also conduct a mechanistic analysis of the learned weights, which constitute an alternative constructive proof that removing causal attention masking admits a parameter configuration which solves the inverse permutation learning problem. 

In \Cref{sec:possibility_copy} we show, more surprisingly, that \emph{padding the input} with additional ``scratch tokens" also yields a construction for inverse permutation learning. As our construction illustrates, this transformation of the input provides the model with ``scratch space'' which can be used to perform additional computation 
during the forward pass. We conjecture that this mechanism may partially explain the success of ``chain-of-thought'' prompting \cite{wei2022cot}, scratchpad prompting \cite{nye2021scratchpad} or the use of ``reasoning tokens'' \cite{deepseek2025r1}, which perform a similar function by allowing the model to perform additional computation \emph{whether or not the intermediate tokens encode meaningful semantic information about the problem}  (e.g., the output of intermediate reasoning steps). This suggests a concrete and, to our knowledge, unexplored mechanism for enabling reasoning in transformer models.\footnote{Perhaps the closest related work is that of \citet{abbe2024reasoning}, who propose the notion of an ``agnostic" (i.e., unsupervised) scratchpad. In this setting, the model may first produce a fixed number of output tokens which do not appear in the training loss for an otherwise standard supervised learning problem. \citet{abbe2024reasoning} conjecture that agnostic scratchpads are incompatible with sample efficient learning. Our results instead suggest padding the \emph{input} with scratch tokens, but can nonetheless be interpreted as evidence which contradicts this conjecture, at least in settings where scratch tokens can help circumvent the constraints imposed by causal attention masking.}

\textbf{Limitations.} We study the stylized transformer first proposed by \citet{nichani_how_2024}. Our model preserves the most important features of the architecture, but omits multi-layer perceptron blocks, multi-head attention, and nontrivial positional embeddings, among other complexities. Our results also assume a natural but particular input representation. Furthermore, our constructive results concern the expressive capacity of transformers, but do not examine training dynamics or sample complexity required to achieve these parameter configurations. 

\textbf{Organization.} We present formal statements of our technical results in \Cref{sec:technicalresults}. We discuss additional related work in \Cref{app:additional_related}. We provide empirical results and conduct a mechanistic analysis to corroborate our primary contributions in \Cref{app:empirical}. All proofs are provided in \Cref{app:proofs}.

\section{Statement of Technical Results} \label{sec:technicalresults}

We begin with the definition of the inverse permutation function.
\begin{definition}[Inverse permutation function] \label{permutation-task}
    For fixed $n$, a permutation matrix $P \in \{ 0, 1\}^{n \times n}$ is defined such that $i$th row represents the elementary unit vector $e_{\pi(i)}$ for some permutation $\pi(\cdot)$. For any $n$-row matrix $Y_P$, the inverse permutation function outputs $Y = P^{-1} Y_P$.
\end{definition}

We interpret each row of $Y_P$ as corresponding to the embedding of a single token in the (permuted) input, and each row of $Y$ as corresponding to the embedding of a single token of the (canonical) output. Observe that one function satisfying \Cref{permutation-task} is the trivial linear transformation $(P, Y_P) \mapsto P^\top Y_P$, since $P^{-1} = P^\top$.
As mentioned above, this task is arguably simpler than the more realistic tasks which motivate it: the description of the permutation $P$ is given explicitly, rather than being determined by some function of the permuted input $Y_P$ (e.g., its lexicographic index).
Thus, we have eliminated the difficulty of determining the order in which inputs are arranged or the choice of canonical ordering during internal model computations.
Similarly, by presenting $Y_P$ as a matrix, we have also reduced the difficulty of the problem by providing a fixed delimiting of the set of examples (rather than the model having to delimit free text into the set of choices).

The decoder-only transformer architecture underpins the most popular frontier LLMs, including the GPT series.\footnote{The most popular open source models (e.g., the Llama series, Qwen series and Mixtral series) are built on decoder-only architectures, as was OpenAI's GPT-3 model. While the architecture for newer closed-source models like GPT-5 or the Gemini, Claude or Grok series has not been publicly disclosed, folklore in the AI research community suggests that these models are also underpinned by variants of a decoder-only architecture.} We focus on a simplified decoder-only, ``disentangled'' transformer architecture proposed by \citet{nichani_how_2024}. We describe this architecture below.

\textbf{A simplified transformer model.} The input will consist of a matrix $X \in \R^{T \times d}$ where $T$ is the length of the sequence and $d$ is the input dimension.
The input sequence is first passed through an embedding function, which consists of the concatenation of token and position embeddings.
For simplicity, we will assume the token embeddings are just $X$ itself and position embeddings are one-hot encodings of each of $X$'s row positions, so that the encoding of an input is 
\begin{align*}
    h^{(0)} \defeq [X , I] \in \R^{T \times (d + T)}
\end{align*}
where $I$ is the $T \times T$ identity matrix.
Subsequent layers of the transformer will consist of attention layers, which are each parameterized by a matrix $A$.
Each attention layer consists of a softmax operation $\cS(\cdot)$ composed with a causal attention mask operation $\MASK(\cdot)$:
\begin{align*}
    \attn(h; A) \defeq \cS(\MASK(h A h^\top)) h
\end{align*}
The softmax operation is applied row-wise to a matrix. For a given vector $v$, we let $\cS(v)_i = \exp(v_i)/\sum_{j} \exp(v_j)$.
For a matrix $V$, the causal attention mask just takes the lower-triangular entries of $V$ so that $\MASK(V)_{ij} = V_{ij}$ if $i \geq j$ and $-\infty$ above the diagonal.
We will denote the weight matrix at the $i$th layer as $A^{(i)}$.

Typically, the outputs of a layer are added to its inputs; i.e., the residual stream out of layer $i+1$ is $h^{(i)} + \attn(h^{(i)}; A^{(i+1)})$.
We will instead analyze a \textit{disentangled transformer}, proposed by \citet{nichani_how_2024}, which concatenates the outputs of a layer with its inputs to form the inputs to the next layer:
\begin{align*}
    h^{(i+1)} \defeq [h^{(i)}, \attn(h^{(i)}; A^{(i+1)})].
\end{align*}
The disentangled transformer is exactly as expressive as a vanilla (layer-sum) transformer (see, \citet{nichani_how_2024}, Theorem 3). It is analytically useful because it clarifies the structure of the \textit{residual stream} \citep{elhage2021mathematical}, which describes how transformers may use orthogonal subspaces of layer outputs as communication channels. In particular, while standard transformers store information in orthogonal subspaces of the residual stream, disentangled transformers store the outputs of each layer in separate matrix blocks.
We will refer to $h^{(i)}$ as the residual stream at layer $i$.
Throughout this paper, we will analyze disentangled transformers.
For simplicity, we will consider transformers with a single head. All of our results hold for multi-head transformers.

For expressibility of inverse permutation functions (\cref{permutation-task}), the input to the transformer is the concatenation of the task inputs, represented as $X = [P; Y_P]$.
We note that $Y_P$ is represented with $d$ rows since this is the dimension of the permutation matrix.
Thus, $T = 2d$.

\subsection{The impossibility of inverse permutation learning with a causal attention mask.} \label{sec:impossibility}

Our main theoretical result states that, for any nontrivial permutation $P$, no decoder-only transformer of any depth can output $Y$ to a block of the residual stream. We state this result below.
\begin{restatable}{theorem}{noundoperms} \label{clm:no-undo-perms}
    For all $k$ and parameter matrices $\{ A^{(i)} \}_{i \in [k]}$, and all permutation matrices $P$ other than the identity permutation, there exists a target matrix $Y$ such that a decoder-only, attention-only transformer parameterized by $\{ A^{(i)}\}_{i \in [k]}$ given $[P; Y_P]$ as input does not output $Y$ to any block of the residual stream.
\end{restatable}

As described above, the intuition for this result is straightforward: for any nontrivial permutation, at least one element which must be moved from a later position to an earlier position. However, the causal attention mask precludes this: any row in the residual stream corresponding to position $i$ must be invariant to changes in rows corresponding to positions $j > i$. This interpretation is supported by \Cref{prop:without_ca}, which we turn to next. A proof of \Cref{clm:no-undo-perms} is provided in \Cref{app:proofs}. 
We note that this theorem does not depend on the order in which $P$ and $Y_P$ are given as input: the same impossibility result holds for reversed inputs $[Y_P; P]$.

\subsection{The possibility of inverse permutation learning with an unrestricted attention mask.}
\label{sec:possibility_nocausal}

Our next result serves to reinforce the key intuition from the prior section; namely, that the causal attention mask precludes the transfer of information which is necessary to perform inverse permutation learning. First, we define the causal mask-free attention operation.

\begin{definition}
    We will say an attention layer is \textit{causal mask-free} (CMF) if it computes
    \begin{align*}
        \attn_{\mathrm{CMF}}(h; A) \defeq \cS(h A h^\top) h.
    \end{align*}
    We say a decoder-only, attention-only transformer is CMF if all of its attention layers are CMF.
\end{definition}
That is, $\attn_{\mathrm{CMF}}$ is defined identically to $\attn$ except that the $\MASK$ operation is removed. In that sense, the causal mask-free attention operation is no longer ``causal'', and instead allows for unrestricted information flow between tokens.

Our next result shows that this change is sufficient to recover a construction which solves the inverse permutation learning task.

\begin{restatable}{theorem}{withoutca} \label{prop:without_ca}
    There exists parameter matrices $A^{(1)}$ and $A^{(2)}$ such that, for any permutation matrix $P$ and target matrix $Y$, a two-layer, decoder-only, attention-only, causal mask-free, transformer parameterized by $A^{(1)}, A^{(2)}$ given $[P; Y_P]$ as input outputs $Y$ to a block of the residual stream.
\end{restatable}

We provide a constructive proof of this result in \Cref{app:proofs}. This result is perhaps unsurprising, and reinforces the interpretation of \Cref{clm:no-undo-perms} by showing that removing the causal attention mask suffices to circumvent this impossibility result. 

We provide complementary empirical results in \Cref{app:empirical}, which demonstrate that performance improves from the level of random guessing to near perfect accuracy if we remove the causal attention mask. We also conduct a mechanistic analysis of the trained model which demonstrates that the learned weights can constitute a constructive proof of \Cref{prop:without_ca}. We now turn to a second construction for inverse permutation learning.

\subsection{The possibility of inverse permutation learning with input padding.}
\label{sec:possibility_copy}

We now turn to a more surprising result --- simply padding the input is also sufficient to demonstrate a construction that solves the inverse permutation learning task. 
Let $s \in \R^d$ be the embedding associated with a special scratch token.
Let $S = \1 s^\top \in \R^{d \times d}$ be the matrix with $d$ rows where each row is $s$.
We will let $s = 0_{1 \times d}$ by convention.
%
%
We state this theorem below.

\begin{restatable}{theorem}{copyinputs}\label{prop:copy_inputs}
    There exist parameter matrices $A^{(1)}$ and $A^{(2)}$ such that, for any permutation matrix $P$ and target matrix $Y$, a two-layer decoder-only attention-only transformer parameterized by $A^{(1)}, A^{(2)}$ given $[P; Y_P; S]$ as input outputs $Y$ to a block of the residual stream.
\end{restatable}
\newcommand{\BOS}{\texttt{BOS}}
We provide a proof in \Cref{app:proofs}. The proof requires that the first token of the input be a special $\BOS$ token with both the token and positional embedding set to all zeros. The proof highlights that the specific content of the ``padding'' tokens (here, the $S$) is unimportant. Instead, as discussed in \Cref{sec:intro}, these padding tokens provide the model with additional ``scratch space'' to perform the necessary matrix operations. Unlike \Cref{prop:without_ca}, this result provides a recipe for inverse permutation learning without modifying the decoder-only architecture which underpins most modern LLMs.

\bibliographystyle{plainnat}
\bibliography{references,zotero}

\begin{thebibliography}{34}
\providecommand{\natexlab}[1]{#1}
\providecommand{\url}[1]{\texttt{#1}}
\expandafter\ifx\csname urlstyle\endcsname\relax
  \providecommand{\doi}[1]{doi: #1}\else
  \providecommand{\doi}{doi: \begingroup \urlstyle{rm}\Url}\fi

\bibitem[Abbe et~al.(2024)Abbe, Bengio, Lotfi, Sandon, and Saremi]{abbe2024reasoning}
Emmanuel Abbe, Samy Bengio, Aryo Lotfi, Colin Sandon, and Omid Saremi.
\newblock How far can transformers reason? the globality barrier and inductive scratchpad, 2024.

\bibitem[Amiri et~al.(2025)Amiri, Huang, Rofin, and Hahn]{amiri2025lower}
Alireza Amiri, Xinting Huang, Mark Rofin, and Michael Hahn.
\newblock Lower bounds for chain-of-thought reasoning in hard-attention transformers, 2025.

\bibitem[Bhope et~al.(2025{\natexlab{a}})Bhope, Venkateswaran, Jayaram, Isahagian, Muthusamy, and Venkatasubramanian]{atul2025opti}
Rahul~Atul Bhope, Praveen Venkateswaran, K.~R. Jayaram, Vatche Isahagian, Vinod Muthusamy, and Nalini Venkatasubramanian.
\newblock Optiseq: Ordering examples on-the-fly for in-context learning, 2025{\natexlab{a}}.

\bibitem[Bhope et~al.(2025{\natexlab{b}})Bhope, Venkateswaran, Jayaram, Isahagian, Muthusamy, Venkatasubramanian, Shin, Razeghi, Logan, Wallace, 2020, Socher, Perelygin, Wu, Chuang, Manning, Ng, Potts, Recursive, Sorensen, Robinson, Rytting, Shaw, Rogers, Delorey, Khalil, Fulda, 2022, An, Touvron, Lavril, Izacard, Martinet, Lachaux, Lacroix, Rozi{\`e}re, Goyal, and Hambro]{Bhope2025OptiSeqOE}
Rahul~Atul Bhope, Praveen Venkateswaran, K.~R. Jayaram, Vatche Isahagian, Vinod Muthusamy, Nalini Venkatasubramanian, Taylor Shin, Yasaman Razeghi, Robert~L Logan, Eric Wallace, Sameer~Singh. 2020, Richard Socher, Alex Perelygin, Jean Wu, Jason Chuang, Christopher~D. Manning, Andrew~Y Ng, Christopher Potts, Recursive, Taylor Sorensen, Joshua Robinson, Christopher Rytting, Alexander~Glenn Shaw, Kyle~Jeffrey Rogers, Alexia~Pauline Delorey, Mahmoud Khalil, Nancy Fulda, David~Wingate 2022, An, Hugo Touvron, Thibaut Lavril, Gautier Izacard, Xavier Martinet, Marie-Anne Lachaux, Timoth{\'e}e Lacroix, Baptiste Rozi{\`e}re, Naman Goyal, and Eric Hambro.
\newblock Optiseq: Ordering examples on-the-fly for in-context learning.
\newblock 2025{\natexlab{b}}.
\newblock URL \url{https://api.semanticscholar.org/CorpusID:275921280}.

\bibitem[Brown et~al.(2020)Brown, Mann, Ryder, Subbiah, Kaplan, Dhariwal, Neelakantan, Shyam, Sastry, Askell, Agarwal, Herbert-Voss, Krueger, Henighan, Child, Ramesh, Ziegler, Wu, Winter, Hesse, Chen, Sigler, Litwin, Gray, Chess, Clark, Berner, McCandlish, Radford, Sutskever, and Amodei]{brown2020fewshot}
Tom~B. Brown, Benjamin Mann, Nick Ryder, Melanie Subbiah, Jared Kaplan, Prafulla Dhariwal, Arvind Neelakantan, Pranav Shyam, Girish Sastry, Amanda Askell, Sandhini Agarwal, Ariel Herbert-Voss, Gretchen Krueger, Tom Henighan, Rewon Child, Aditya Ramesh, Daniel~M. Ziegler, Jeffrey Wu, Clemens Winter, Christopher Hesse, Mark Chen, Eric Sigler, Mateusz Litwin, Scott Gray, Benjamin Chess, Jack Clark, Christopher Berner, Sam McCandlish, Alec Radford, Ilya Sutskever, and Dario Amodei.
\newblock Language models are few-shot learners, 2020.

\bibitem[Chen et~al.(2024)Chen, Chi, Wang, and Zhou]{Chen2024-ox}
Xinyun Chen, Ryan~A Chi, Xuezhi Wang, and Denny Zhou.
\newblock Premise order matters in reasoning with large language models.
\newblock February 2024.

\bibitem[Cohen-Karlik et~al.(2020)Cohen-Karlik, David, and Globerson]{karlik2020invariance}
Edo Cohen-Karlik, Avichai~Ben David, and Amir Globerson.
\newblock Regularizing towards permutation invariance in recurrent models, 2020.

\bibitem[DeepSeek-AI et~al.(2025)DeepSeek-AI, Guo, Yang, Zhang, Song, Zhang, Xu, Zhu, Ma, Wang, Bi, Zhang, Yu, Wu, Wu, Gou, Shao, Li, Gao, Liu, Xue, Wang, Wu, Feng, Lu, Zhao, Deng, Zhang, Ruan, Dai, Chen, Ji, Li, Lin, Dai, Luo, Hao, Chen, Li, Zhang, Bao, Xu, Wang, Ding, Xin, Gao, Qu, Li, Guo, Li, Wang, Chen, Yuan, Qiu, Li, Cai, Ni, Liang, Chen, Dong, Hu, Gao, Guan, Huang, Yu, Wang, Zhang, Zhao, Wang, Zhang, Xu, Xia, Zhang, Zhang, Tang, Li, Wang, Li, Tian, Huang, Zhang, Wang, Chen, Du, Ge, Zhang, Pan, Wang, Chen, Jin, Chen, Lu, Zhou, Chen, Ye, Wang, Yu, Zhou, Pan, Li, Zhou, Wu, Ye, Yun, Pei, Sun, Wang, Zeng, Zhao, Liu, Liang, Gao, Yu, Zhang, Xiao, An, Liu, Wang, Chen, Nie, Cheng, Liu, Xie, Liu, Yang, Li, Su, Lin, Li, Jin, Shen, Chen, Sun, Wang, Song, Zhou, Wang, Shan, Li, Wang, Wei, Zhang, Xu, Li, Zhao, Sun, Wang, Yu, Zhang, Shi, Xiong, He, Piao, Wang, Tan, Ma, Liu, Guo, Ou, Wang, Gong, Zou, He, Xiong, Luo, You, Liu, Zhou, Zhu, Xu, Huang, Li, Zheng, Zhu, Ma, Tang, Zha, Yan, Ren, Ren, Sha, Fu, Xu, Xie, Zhang,
  Hao, Ma, Yan, Wu, Gu, Zhu, Liu, Li, Xie, Song, Pan, Huang, Xu, Zhang, and Zhang]{deepseek2025r1}
DeepSeek-AI, Daya Guo, Dejian Yang, Haowei Zhang, Junxiao Song, Ruoyu Zhang, Runxin Xu, Qihao Zhu, Shirong Ma, Peiyi Wang, Xiao Bi, Xiaokang Zhang, Xingkai Yu, Yu~Wu, Z.~F. Wu, Zhibin Gou, Zhihong Shao, Zhuoshu Li, Ziyi Gao, Aixin Liu, Bing Xue, Bingxuan Wang, Bochao Wu, Bei Feng, Chengda Lu, Chenggang Zhao, Chengqi Deng, Chenyu Zhang, Chong Ruan, Damai Dai, Deli Chen, Dongjie Ji, Erhang Li, Fangyun Lin, Fucong Dai, Fuli Luo, Guangbo Hao, Guanting Chen, Guowei Li, H.~Zhang, Han Bao, Hanwei Xu, Haocheng Wang, Honghui Ding, Huajian Xin, Huazuo Gao, Hui Qu, Hui Li, Jianzhong Guo, Jiashi Li, Jiawei Wang, Jingchang Chen, Jingyang Yuan, Junjie Qiu, Junlong Li, J.~L. Cai, Jiaqi Ni, Jian Liang, Jin Chen, Kai Dong, Kai Hu, Kaige Gao, Kang Guan, Kexin Huang, Kuai Yu, Lean Wang, Lecong Zhang, Liang Zhao, Litong Wang, Liyue Zhang, Lei Xu, Leyi Xia, Mingchuan Zhang, Minghua Zhang, Minghui Tang, Meng Li, Miaojun Wang, Mingming Li, Ning Tian, Panpan Huang, Peng Zhang, Qiancheng Wang, Qinyu Chen, Qiushi Du, Ruiqi Ge, Ruisong
  Zhang, Ruizhe Pan, Runji Wang, R.~J. Chen, R.~L. Jin, Ruyi Chen, Shanghao Lu, Shangyan Zhou, Shanhuang Chen, Shengfeng Ye, Shiyu Wang, Shuiping Yu, Shunfeng Zhou, Shuting Pan, S.~S. Li, Shuang Zhou, Shaoqing Wu, Shengfeng Ye, Tao Yun, Tian Pei, Tianyu Sun, T.~Wang, Wangding Zeng, Wanjia Zhao, Wen Liu, Wenfeng Liang, Wenjun Gao, Wenqin Yu, Wentao Zhang, W.~L. Xiao, Wei An, Xiaodong Liu, Xiaohan Wang, Xiaokang Chen, Xiaotao Nie, Xin Cheng, Xin Liu, Xin Xie, Xingchao Liu, Xinyu Yang, Xinyuan Li, Xuecheng Su, Xuheng Lin, X.~Q. Li, Xiangyue Jin, Xiaojin Shen, Xiaosha Chen, Xiaowen Sun, Xiaoxiang Wang, Xinnan Song, Xinyi Zhou, Xianzu Wang, Xinxia Shan, Y.~K. Li, Y.~Q. Wang, Y.~X. Wei, Yang Zhang, Yanhong Xu, Yao Li, Yao Zhao, Yaofeng Sun, Yaohui Wang, Yi~Yu, Yichao Zhang, Yifan Shi, Yiliang Xiong, Ying He, Yishi Piao, Yisong Wang, Yixuan Tan, Yiyang Ma, Yiyuan Liu, Yongqiang Guo, Yuan Ou, Yuduan Wang, Yue Gong, Yuheng Zou, Yujia He, Yunfan Xiong, Yuxiang Luo, Yuxiang You, Yuxuan Liu, Yuyang Zhou, Y.~X. Zhu,
  Yanhong Xu, Yanping Huang, Yaohui Li, Yi~Zheng, Yuchen Zhu, Yunxian Ma, Ying Tang, Yukun Zha, Yuting Yan, Z.~Z. Ren, Zehui Ren, Zhangli Sha, Zhe Fu, Zhean Xu, Zhenda Xie, Zhengyan Zhang, Zhewen Hao, Zhicheng Ma, Zhigang Yan, Zhiyu Wu, Zihui Gu, Zijia Zhu, Zijun Liu, Zilin Li, Ziwei Xie, Ziyang Song, Zizheng Pan, Zhen Huang, Zhipeng Xu, Zhongyu Zhang, and Zhen Zhang.
\newblock Deepseek-r1: Incentivizing reasoning capability in llms via reinforcement learning, 2025.

\bibitem[Egressy and St{\"u}hmer(2025)]{Egressy2025-zv}
Beni Egressy and Jan St{\"u}hmer.
\newblock {Set-LLM}: A {Permutation-Invariant} {LLM}.
\newblock May 2025.

\bibitem[Elhage et~al.(2021)Elhage, Nanda, Olsson, Henighan, Joseph, Mann, Askell, Bai, Chen, Conerly, et~al.]{elhage2021mathematical}
Nelson Elhage, Neel Nanda, Catherine Olsson, Tom Henighan, Nicholas Joseph, Ben Mann, Amanda Askell, Yuntao Bai, Anna Chen, Tom Conerly, et~al.
\newblock A mathematical framework for transformer circuits.
\newblock \emph{Transformer Circuits Thread}, 1\penalty0 (1):\penalty0 12, 2021.

\bibitem[Gardner et~al.(2024)Gardner, Perdomo, and Schmidt]{gardner2024tabula}
Josh Gardner, Juan~C. Perdomo, and Ludwig Schmidt.
\newblock Large scale transfer learning for tabular data via language modeling, 2024.

\bibitem[Giapitzakis et~al.(2025)Giapitzakis, de~Luca, and Fountoulakis]{giapitzakis2025learning}
George Giapitzakis, Artur~Back de~Luca, and Kimon Fountoulakis.
\newblock Learning to add, multiply, and execute algorithmic instructions exactly with neural networks.
\newblock \emph{arXiv preprint arXiv:2502.16763}, 2025.

\bibitem[Guo et~al.(2024)Guo, Wang, Wang, Ye, and Zhang]{guo-etal-2024-makes}
Qi~Guo, Leiyu Wang, Yidong Wang, Wei Ye, and Shikun Zhang.
\newblock What makes a good order of examples in in-context learning.
\newblock In Lun-Wei Ku, Andre Martins, and Vivek Srikumar, editors, \emph{Findings of the Association for Computational Linguistics: ACL 2024}, pages 14892--14904, Bangkok, Thailand, August 2024. Association for Computational Linguistics.
\newblock \doi{10.18653/v1/2024.findings-acl.884}.
\newblock URL \url{https://aclanthology.org/2024.findings-acl.884/}.

\bibitem[Han et~al.(2022)Han, Schoelkopf, Zhao, Qi, Riddell, Zhou, Coady, Peng, Qiao, Benson, Sun, Wardle-Solano, Szabo, Zubova, Burtell, Fan, Liu, Wong, Sailor, Ni, Nan, Kasai, Yu, Zhang, Fabbri, Kryscinski, Yavuz, Liu, Lin, Joty, Zhou, Xiong, Ying, Cohan, and Radev]{Han2022-hm}
Simeng Han, Hailey Schoelkopf, Yilun Zhao, Zhenting Qi, Martin Riddell, Wenfei Zhou, James Coady, David Peng, Yujie Qiao, Luke Benson, Lucy Sun, Alex Wardle-Solano, Hannah Szabo, Ekaterina Zubova, Matthew Burtell, Jonathan Fan, Yixin Liu, Brian Wong, Malcolm Sailor, Ansong Ni, Linyong Nan, Jungo Kasai, Tao Yu, Rui Zhang, Alexander~R Fabbri, Wojciech Kryscinski, Semih Yavuz, Ye~Liu, Xi~Victoria Lin, Shafiq Joty, Yingbo Zhou, Caiming Xiong, Rex Ying, Arman Cohan, and Dragomir Radev.
\newblock {FOLIO}: Natural language reasoning with first-order logic.
\newblock September 2022.

\bibitem[Hendrycks et~al.(2020)Hendrycks, Burns, Basart, Zou, Mazeika, Song, and Steinhardt]{Hendrycks2020-uh}
Dan Hendrycks, Collin Burns, Steven Basart, Andy Zou, Mantas Mazeika, Dawn Song, and Jacob Steinhardt.
\newblock Measuring massive multitask language understanding.
\newblock September 2020.

\bibitem[Hollmann et~al.(2022)Hollmann, Müller, Eggensperger, and Hutter]{hollmann20222tabpfn}
Noah Hollmann, Samuel Müller, Katharina Eggensperger, and Frank Hutter.
\newblock Tabpfn: A transformer that solves small tabular classification problems in a second, 2022.

\bibitem[Li et~al.(2025)Li, Guo, and Andreas]{li2025language}
Belinda~Z Li, Zifan~Carl Guo, and Jacob Andreas.
\newblock (how) do language models track state?
\newblock \emph{arXiv preprint arXiv:2503.02854}, 2025.

\bibitem[Li et~al.(2022)Li, Hopkins, Bau, Vi{\'e}gas, Pfister, and Wattenberg]{li2022emergent}
Kenneth Li, Aspen~K Hopkins, David Bau, Fernanda Vi{\'e}gas, Hanspeter Pfister, and Martin Wattenberg.
\newblock Emergent world representations: Exploring a sequence model trained on a synthetic task. arxiv.
\newblock \emph{arXiv preprint arXiv:2210.13382}, 2022.

\bibitem[Li and Qiu(2023)]{Li2023-kk}
Xiaonan Li and Xipeng Qiu.
\newblock Finding supporting examples for in-context learning.
\newblock February 2023.

\bibitem[Liu et~al.(2023{\natexlab{a}})Liu, Lin, Hewitt, Paranjape, Bevilacqua, Petroni, and Liang]{Liu2023-nb}
Nelson~F Liu, Kevin Lin, John Hewitt, Ashwin Paranjape, Michele Bevilacqua, Fabio Petroni, and Percy Liang.
\newblock Lost in the middle: How language models use long contexts.
\newblock July 2023{\natexlab{a}}.

\bibitem[Liu et~al.(2023{\natexlab{b}})Liu, Lin, Hewitt, Paranjape, Bevilacqua, Petroni, and Liang]{Liu2023-ot}
Nelson~F Liu, Kevin Lin, John Hewitt, Ashwin Paranjape, Michele Bevilacqua, Fabio Petroni, and Percy Liang.
\newblock Lost in the middle: How language models use long contexts.
\newblock July 2023{\natexlab{b}}.

\bibitem[Lu et~al.(2021)Lu, Bartolo, Moore, Riedel, and Stenetorp]{Lu2021-lz}
Yao Lu, Max Bartolo, Alastair Moore, Sebastian Riedel, and Pontus Stenetorp.
\newblock Fantastically ordered prompts and where to find them: Overcoming few-shot prompt order sensitivity.
\newblock April 2021.

\bibitem[Nichani et~al.(2024)Nichani, Damian, and Lee]{nichani_how_2024}
Eshaan Nichani, Alex Damian, and Jason~D. Lee.
\newblock How {Transformers} {Learn} {Causal} {Structure} with {Gradient} {Descent}, August 2024.
\newblock URL \url{http://arxiv.org/abs/2402.14735}.
\newblock arXiv:2402.14735 [cs].

\bibitem[Nye et~al.(2021)Nye, Andreassen, Gur-Ari, Michalewski, Austin, Bieber, Dohan, Lewkowycz, Bosma, Luan, Sutton, and Odena]{nye2021scratchpad}
Maxwell Nye, Anders~Johan Andreassen, Guy Gur-Ari, Henryk Michalewski, Jacob Austin, David Bieber, David Dohan, Aitor Lewkowycz, Maarten Bosma, David Luan, Charles Sutton, and Augustus Odena.
\newblock Show your work: Scratchpads for intermediate computation with language models, 2021.

\bibitem[Pezeshkpour and Hruschka(2023)]{Pezeshkpour2023-ct}
Pouya Pezeshkpour and Estevam Hruschka.
\newblock Large language models sensitivity to the order of options in multiple-choice questions.
\newblock August 2023.

\bibitem[Richens et~al.(2025)Richens, Abel, Bellot, and Everitt]{richens2025general}
Jonathan Richens, David Abel, Alexis Bellot, and Tom Everitt.
\newblock General agents need world models.
\newblock \emph{arXiv preprint arXiv:2506.01622}, 2025.

\bibitem[Shi et~al.(2024)Shi, Ma, Liang, Diao, Ma, and Vosoughi]{Shi2024-ar}
Lin Shi, Chiyu Ma, Wenhua Liang, Xingjian Diao, Weicheng Ma, and Soroush Vosoughi.
\newblock Judging the judges: A systematic study of position bias in {LLM-as-a-Judge}.
\newblock June 2024.

\bibitem[Toshniwal et~al.(2022)Toshniwal, Wiseman, Livescu, and Gimpel]{toshniwal2022chess}
Shubham Toshniwal, Sam Wiseman, Karen Livescu, and Kevin Gimpel.
\newblock Chess as a testbed for language model state tracking.
\newblock In \emph{Proceedings of the AAAI Conference on Artificial Intelligence}, volume~36, pages 11385--11393, 2022.

\bibitem[Vafa et~al.(2024)Vafa, Chen, Kleinberg, Mullainathan, and Rambachan]{vafa_evaluating_2024}
Keyon Vafa, Justin~Y. Chen, Jon Kleinberg, Sendhil Mullainathan, and Ashesh Rambachan.
\newblock Evaluating the {World} {Model} {Implicit} in a {Generative} {Model}, June 2024.
\newblock URL \url{http://arxiv.org/abs/2406.03689}.
\newblock arXiv:2406.03689 [cs].

\bibitem[Vafa et~al.(2025)Vafa, Chang, Rambachan, and Mullainathan]{vafa2025has}
Keyon Vafa, Peter~G Chang, Ashesh Rambachan, and Sendhil Mullainathan.
\newblock What has a foundation model found? using inductive bias to probe for world models.
\newblock \emph{arXiv preprint arXiv:2507.06952}, 2025.

\bibitem[Vaswani et~al.(2017)Vaswani, Shazeer, Parmar, Uszkoreit, Jones, Gomez, Kaiser, and Polosukhin]{vaswani2017attention}
Ashish Vaswani, Noam Shazeer, Niki Parmar, Jakob Uszkoreit, Llion Jones, Aidan~N Gomez, {\L}ukasz Kaiser, and Illia Polosukhin.
\newblock Attention is all you need.
\newblock \emph{Advances in neural information processing systems}, 30, 2017.

\bibitem[Wei et~al.(2022)Wei, Wang, Schuurmans, Bosma, Ichter, Xia, Chi, Le, and Zhou]{wei2022cot}
Jason Wei, Xuezhi Wang, Dale Schuurmans, Maarten Bosma, Brian Ichter, Fei Xia, Ed~Chi, Quoc Le, and Denny Zhou.
\newblock Chain-of-thought prompting elicits reasoning in large language models, 2022.

\bibitem[Wen et~al.(2024)Wen, Zhang, Lin, and Zhang]{wen2024cot}
Kaiyue Wen, Huaqing Zhang, Hongzhou Lin, and Jingzhao Zhang.
\newblock From sparse dependence to sparse attention: Unveiling how chain-of-thought enhances transformer sample efficiency, 2024.

\bibitem[Xu et~al.(2023)Xu, Yang, Liu, and He]{Xu2023-zh}
Renjun Xu, Kaifan Yang, Ke~Liu, and Fengxiang He.
\newblock {$E(2)$-Equivariant} vision transformer.
\newblock June 2023.

\end{thebibliography}

\appendix

\section{Additional related work}
\label{app:additional_related}

Our work builds on a rich and rapidly growing literature on reasoning, world modeling and state tracking modern large language models. It is perhaps most closely related to~\citet{li2025language}, which examines whether and how LLMs track state in a permutation composition task. Unlike our work, they explicitly train models to output intermediate states. Another related line of work interrogates the coherence of internal ``world models''~\citep{vafa_evaluating_2024,vafa2025has,li2022emergent,toshniwal2022chess}; these internal models are desirable precisely because they guarantee, for example, that the behavior of the LLM is invariant to perturbations of inputs which correspond to the same logical state. Recent work argues that such a world model is \emph{necessary} for general purpose agents~\citep{richens2025general}. Our work is also closely related to~\citet{giapitzakis2025learning}, which studies the task of learning permutations (given an input, apply a given, fixed permutation to it) in two-layer feedforward neural networks. In contrast, the task we study involves the ``in context'' inversion of arbitrary permutations given as inputs. Our technical results build on the ``disentangled'' transformer model, first proposed by \citet{nichani_how_2024} to study the emergence of causal reasoning in transformer models. Our work also parallels that of \cite{abbe2024reasoning, wen2024cot, amiri2025lower}, who study conditions under which additional output tokens (e.g., chain-of-thought or scratchpad prompting) can improve the expressivity or sample efficiency of transformer models. 

More generally, we build on work studying transformers~\citep{vaswani2017attention} and their capabilities.
In particular, the problem of reconstructing ``canonical'' permutations can be viewed as a building block towards robust in-context learning~\citep{brown2020fewshot} and tabular foundation models models~\citep{hollmann20222tabpfn, gardner2024tabula}, wherein performance should be invariant to permutations of in-context examples or columns.

\section{Empirical validation} \label{app:empirical}

Here we provide details on our empirical validation of our theoretical results. All of our code was forked from the code provided in \citet{nichani_how_2024} (the model training, plotting and logging are all nearly identical to their code; the data generating process and model itself are modified to fit our question and setting). Our code is available at \texttt{\href{https://github.com/johnchrishays/icl/}{https://github.com/johnchrishays/icl/}}.

\subsection{Setup.} We train a disentangled transformer using the architecture described in \Cref{sec:technicalresults}.
The only difference is that, in the main text, our results are about the (possibility or impossibility) of copying $Y$ to the \textit{residual stream} and to train a model, we must produce some \textit{output}.
This is to enable taking gradients and otherwise applying a standard training recipe to the inverse permutation learning problem.
To generate outputs, we define output weights $W$ and multiply the residual stream output in the last layer with $W$: $h^{(2)} W^{\top}$.

We train disentangled transformers with a causal mask and without a causal mask, as defined in \Cref{sec:impossibility} and \Cref{sec:possibility_nocausal}, respectively.
For each of the experiments, the training parameters, loss function and all other details of the implementation and training are exactly the same (including the random seed).

We use squared loss to optimize the model, $2^{16}$ training steps and batches of size 1024.
Model training takes less than 5 minutes on one Nvidia A100 GPU. We set the dimension $d=10$.

To generate the inputs, we generated $P$ by sampling uniformly at random from the set of permutations on $d$ elements.
To generate $Y$, we sampled each entry of a $d \times d$ matrix uniformly at random from $\{0, 1\}$.

\subsection{Results.} 

\paragraph{Transformers with causal masks.} After training the model with causal mask, mean squared error of the model is approximately 2.5, which is the MSE corresponding to random guessing.

\begin{figure}[h]
    \centering
    \begin{subfigure}[b]{0.45\textwidth}
    \centering
        \includegraphics[width=0.8\linewidth]{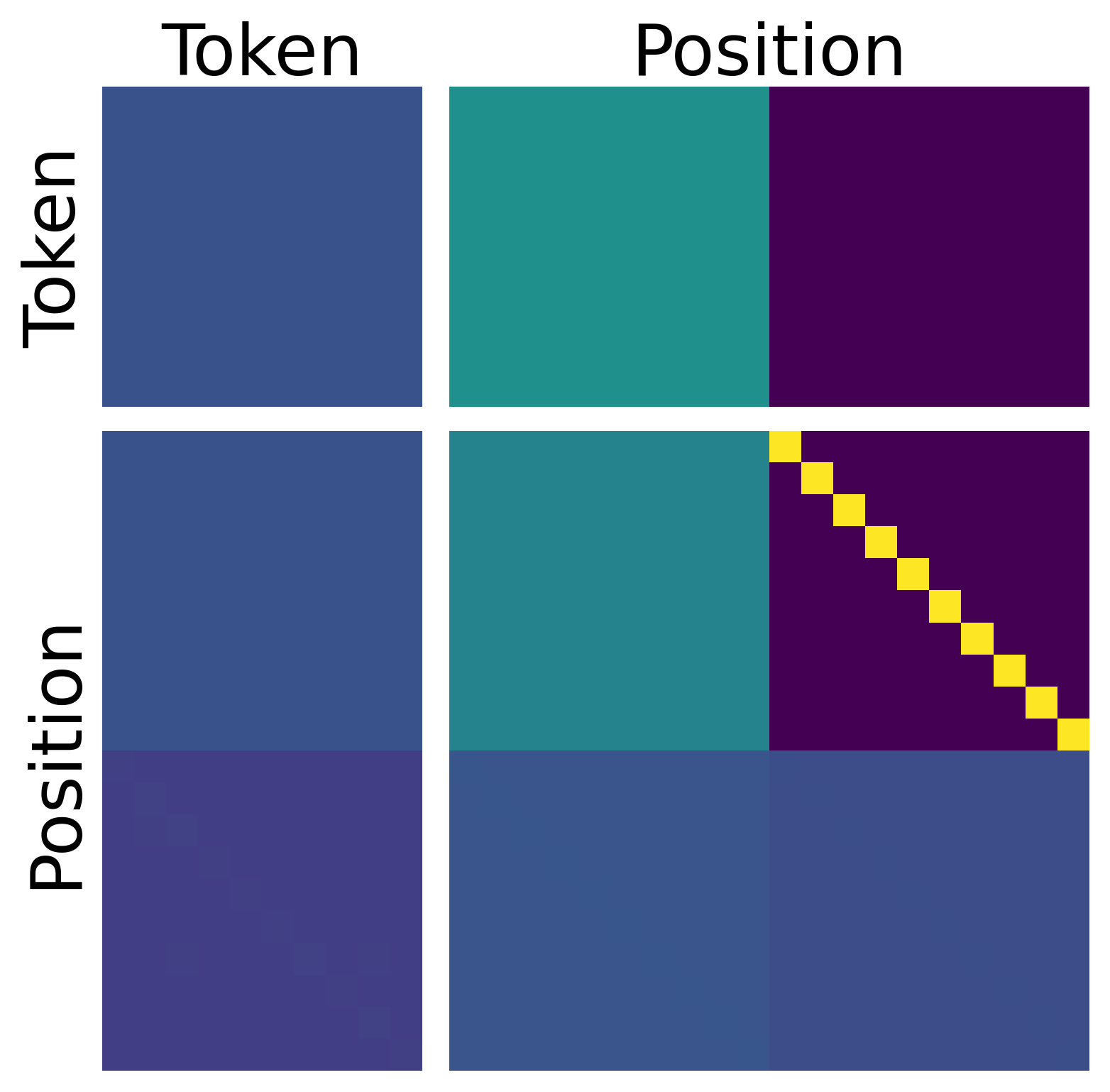}
        \caption{Weights for first attention layer $A^{(1)}$}
        \label{fig:sub-a}
    \end{subfigure}
    \begin{subfigure}[b]{0.45\textwidth}
    \centering
        \includegraphics[width=0.8\linewidth]{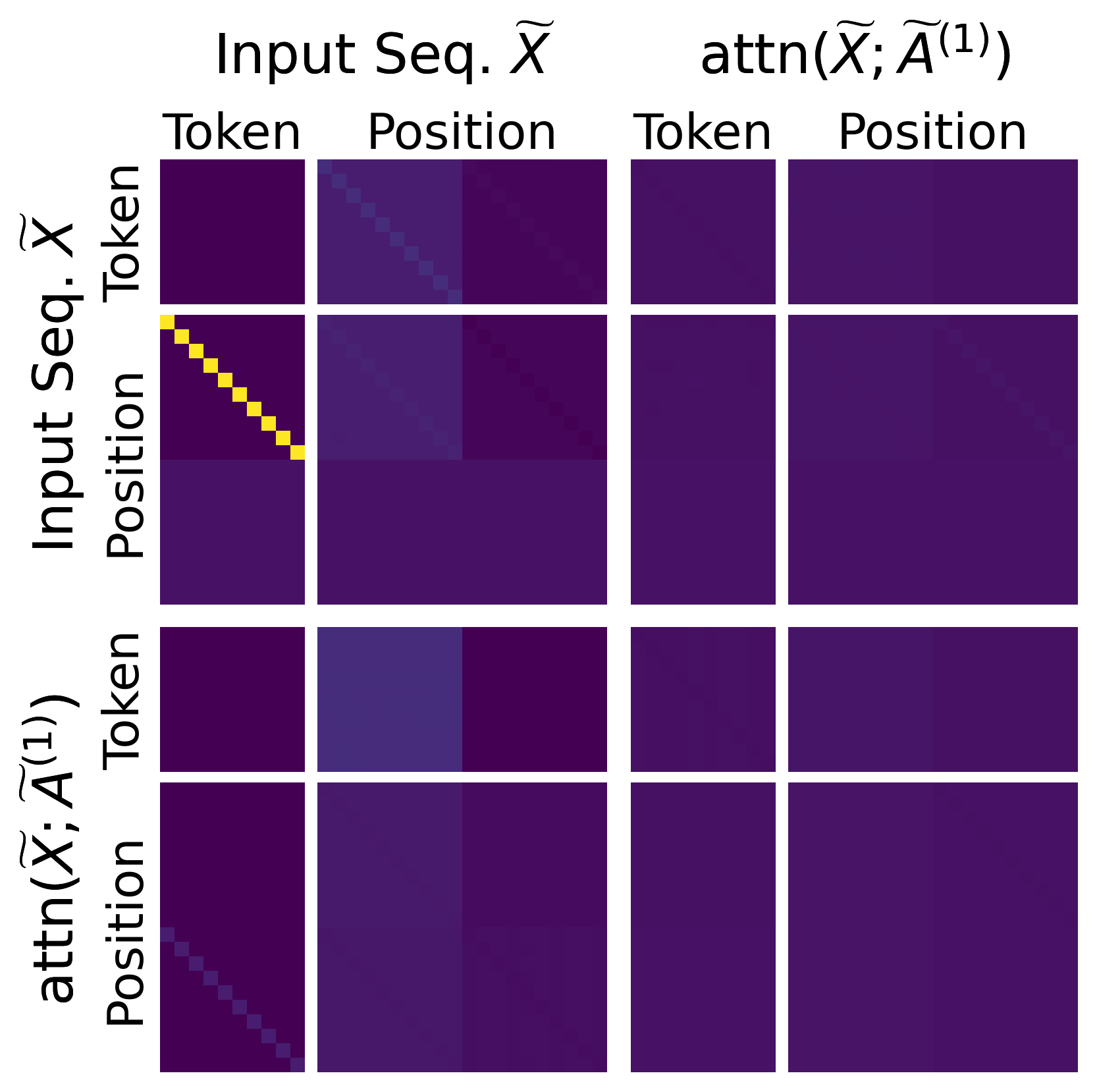}
        \caption{Weights for second attention layer $A^{(2)}$}
    \end{subfigure}
    \begin{subfigure}[b]{\textwidth}
    \centering
        \includegraphics[width=\linewidth]{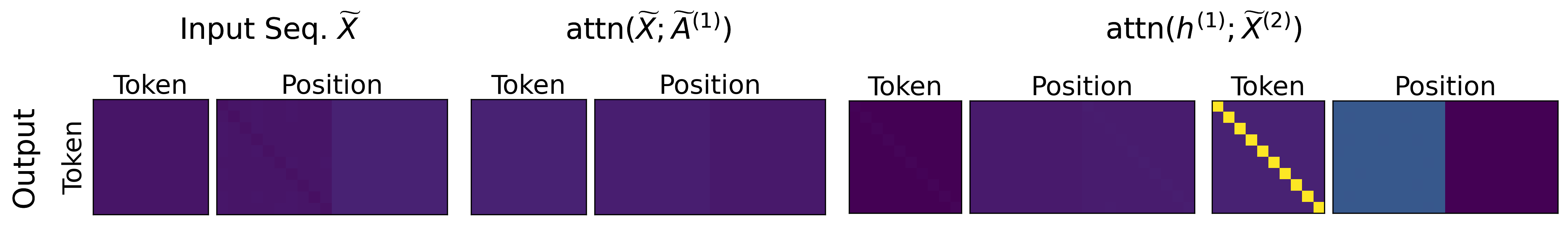}
        \caption{Weights for output layer $W$}
    \end{subfigure}
    \caption{Weights for decoder-only causal-mask-free disentangled transformer trained to invert permutations.}    
    \label{fig:weights}
\end{figure}

\paragraph{Transformers without causal masks.} After training the model without the causal mask, the MSE of the model is about 0.00015. We reproduce the weights corresponding to the trained model in \Cref{fig:weights}.
Panel (a) shows the weights for the first attention layer $A^{(1)}$, panel (b) is for the second attention layer $A^{(2)}$ and (c) is for the output layer $W$.
The construction recovered by the model training process is not the same as that of the construction provided in our proof of \cref{prop:without_ca}; both are valid.
We visualize weights with a heatmap: each small square in each figure represents a single weight parameter.
Thus, $A^{(1)}$ is a matrix of size $3d \times 3d = 30 \times 30$.
In (a), we label weights corresponding to token and position embeddings above and to the side of the matrix.
Weights that are larger are closer to yellow and weights that are close to zero are dark blue.
Thus, the weights matrix $A^{(1)}$ is approximately equal to 
\begin{align*}
    A^{(1)} = \beta \sqparen{\begin{matrix}
        0 & 0 & 0 \\
        0& 0 & I \\
        0 & 0 & 0
    \end{matrix}}
\end{align*}
for some $\beta \gg 0$, where each entry corresponds to a $d \times d$ block. 

In (b), we label weights corresponding to the different blocks of the residual stream: the top/left blocks correspond to the input sequence and the bottom/right blocks correspond to the layer-1 outputs.
Thus, the weights matrix $A^{(2)}$ is approximately equal to
\begin{align*}
    {A}^{(2)} = \beta \sqparen{\begin{array}{ccc|ccc}
        0 & 0 & 0 & 0 & 0 & 0  \\
        I & 0 & 0 & 0& 0 & 0 \\
        0 & 0 & 0 & 0& 0 & 0 \\
        \hline
        0 & 0 & 0 & 0& 0 & 0 \\
        0& 0  & 0 & 0& 0 & 0  \\
        0 & 0 & 0 & 0& 0 & 0 \\
    \end{array}}
\end{align*}
for some $\beta \gg 0$.

In (c), the output weights $W$ can be seen to be approximately equal to the all-zeros matrix except for the 10th block, which is the identity matrix.

These learned weights constitute a constructive proof of \Cref{prop:without_ca}.

\begin{figure}[h]
    \centering
    \begin{subfigure}[b]{0.45\textwidth}
    \centering
        \includegraphics[width=0.8\linewidth]{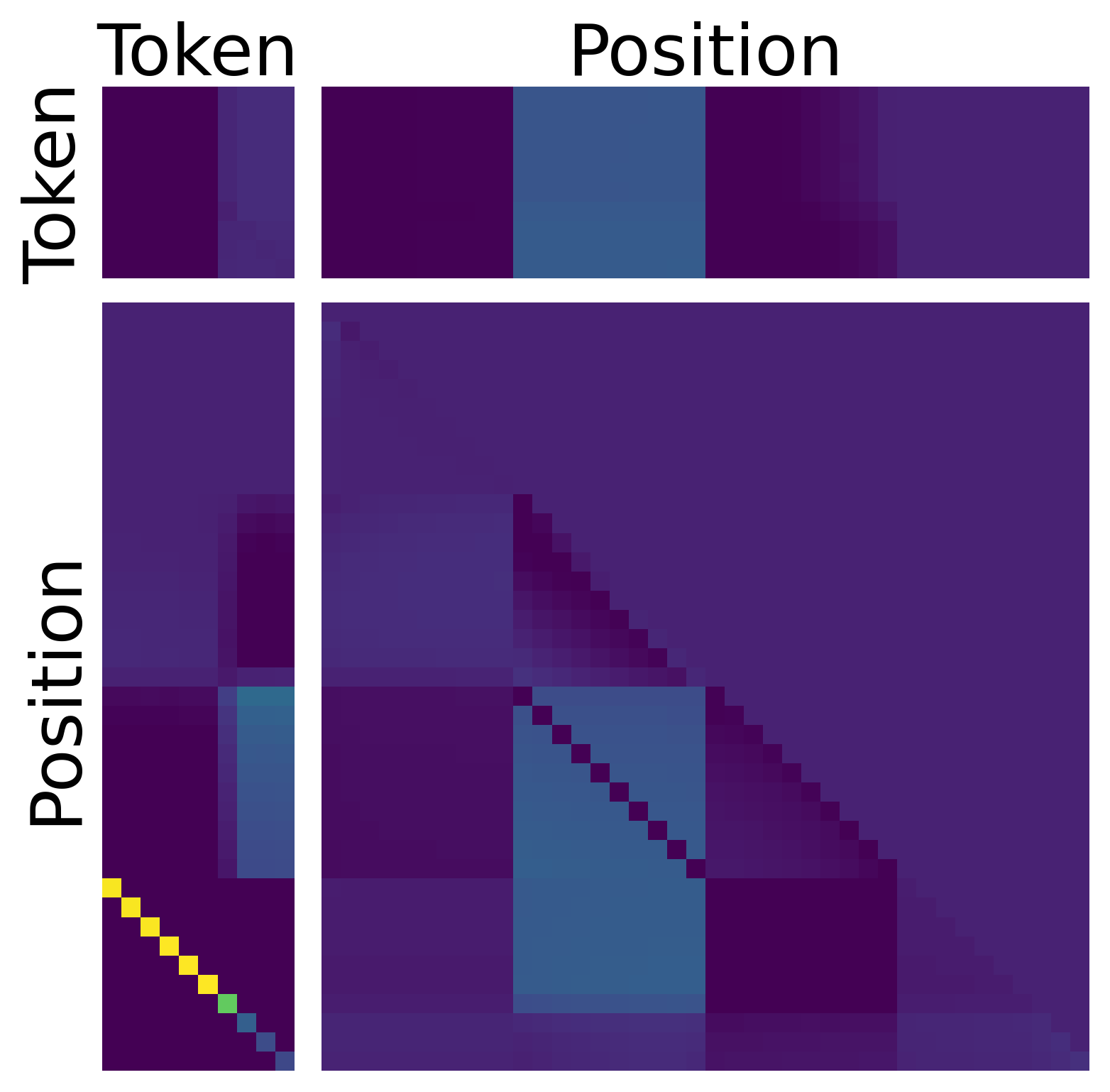}
        \caption{Weights for first attention layer $A^{(1)}$}
        \label{fig:sub-a}
    \end{subfigure}
    \begin{subfigure}[b]{0.45\textwidth}
    \centering
        \includegraphics[width=0.8\linewidth]{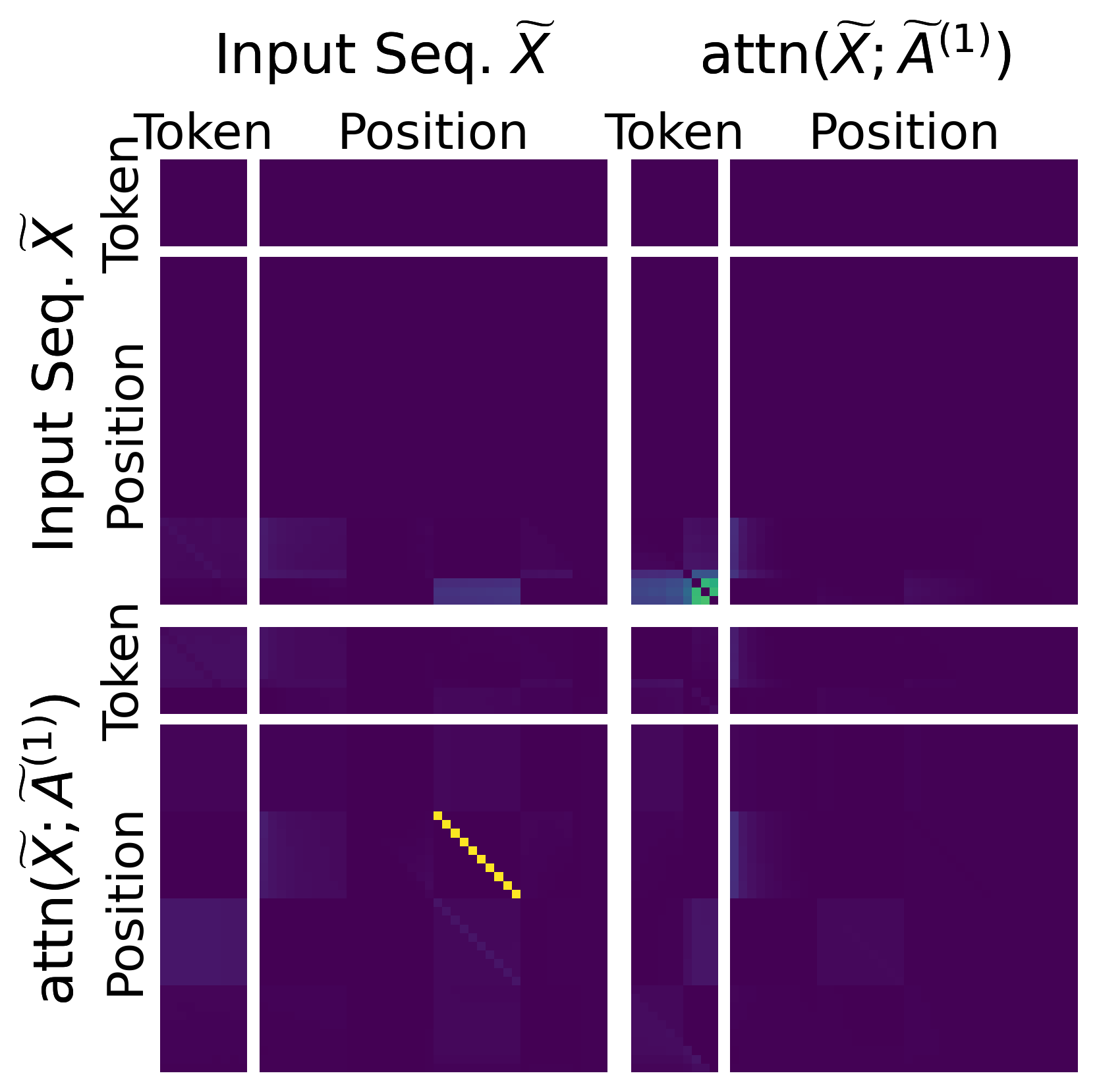}
        \caption{Weights for second attention layer $A^{(2)}$}
    \end{subfigure}
    \begin{subfigure}[b]{\textwidth}
    \centering
        \includegraphics[width=\linewidth]{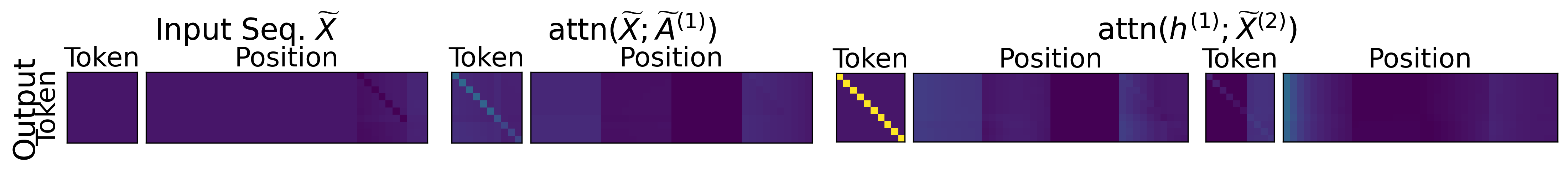}
        \caption{Weights for output layer $W$}
    \end{subfigure}
    \caption{Weights for decoder-only disentangled transformer with scratch space trained to invert permutations.}    
    \label{fig:weights_scratch}
\end{figure}

\paragraph{Transformers with scratch space.} To facilitate ease of visual inspection of the trained transformers, we modify the training setup from above so that there is a $d$-dimensional all-zeroes matrix prepended to the input. 
After training the model, the test MSE is 0.00013. We reproduce the weights corresponding to the trained model in \Cref{fig:weights_scratch}.
The panels display the same quantities as in the case of the causal-mask-free transformer. These weights constitute an alternative constructive proof of Theorem~\ref{prop:copy_inputs}.

\section{Proofs}
\label{app:proofs}

We begin by stating a lemma we will use in the proof of our main theorem.
It formalizes the notion that the causal mask induces an invariance of each row to any subsequent row in the residual stream.
%
%

\begin{lemma} \label{lem:noyptop}
    Let $h$ and $h'$ be any two inputs to the $\ell$th layer such that their first $i$ rows are the same: $h_{1:i} = h'_{1:i}$.
    Then for any weights $A$, $\attn(h; A)_{i} = \attn(h'; A)_{i}$.
\end{lemma}

\begin{proof}
    By inspecting the activation function, we observe the $i$th row of the output of a given layer with weights $A$ are:
    \begin{align*}
        (\activations{h A h^\top} h)_i &= \frac{1}{\sum_{j \leq i} \exp(h_i A h_j^\top)}\sum_{j \leq i} \exp(h_i A h_j^\top) h_j
    \end{align*}
    Since this function does not depend on any row $h_j$ for $j > i$, $\attn(h; A)_{\ell} = \attn(h'; A)_{\ell}$.
\end{proof}

\subsection{Proof of Theorem 1}
\label{app:proof_thm1}
    For ease of reference, we restate each result before its proof.

    \noundoperms*

    As discussed above, the intuition for this result is straightforward: for any permutation other than the identity permutation, there exists at least one element which must be moved from a later position to an earlier position in the sequence. We'll argue, by \Cref{lem:noyptop}, that the causal attention mask precludes this.

    \begin{proof}
    Consider some $P$ that is not the identity matrix.
    Recall that rows and columns of $P$ must sum to one.
    Let $P_i$ denote the $i$th-row of $P$.
    We first prove that $P$ must have at least one nonzero below-diagonal entry.
    That is, we claim that there exists $i$ such that $P_i = e_j$ for $j < i$.
    To see this, let $i$ be the index of the last row which is not equal to its corresponding elementary basis vector $e_i$.
    (By the fact that $P$ is not the identity, there must some such row index.)
    Since for all $j > i$, the $j$th row is equal to $e_j$, the last $d - i$ entries of $P_i$ must be zero.
    Also, since $P_i \neq e_i$, the diagonal entry must be zero.
    Therefore, the non-zero entry of $P_i$ must be at some entry $j$ for $j < i$. 
    This is a non-zero below-diagonal entry of $P_i$.
    Put another way: any non-identity permutation has at least one ``cycle'' ($i_1 \to i_2 \to \dots \to i_1$), which implies there exists a nonzero below-diagonal entry.

    Now, let $i,j$ be any indices such that $P_i = e_j$ for $j < i$.
    In order to store $Y$ in some block of the residual stream, the $j$th row of $Y$ cannot be output into any row of the residual stream more than $d-j$ rows from the bottom. 
    Intuitively, if $Y$ is to ``fit'' into the residual stream, row $j$ cannot be output at too low of a row in the residual stream: there must be space for at least $d-j$ more rows.
    Formally, the $j$th row of $Y$ must be output somewhere in the $\ell$-th row of the residual stream for $\ell \leq d + j + 1$.
    However, since $Y_P = P Y$ and $j < i$, $Y_j$ is located in the $i$th row of $Y_P$, which is in row $d + i + 1$ of the input.
    Since $d + i + 1 > d + j + 1 \geq \ell$, by \cref{lem:noyptop}, row $\ell$ of the residual stream must be invariant to row $d + i + 1$.
    Thus, for any fixed weight matrices $\{ A^{(m)} \}_{m \in [k]}$, there exists some $Y$ such that it is not output to the residual stream.
    \end{proof}

\subsection{Proof of Theorem 2}
\label{app:proof_thm2}

\withoutca*

\begin{proof}[Proof of \Cref{prop:without_ca}] We proceed by construction.

The first layer copies $P$ into the bottom block-row of the residual stream.
For constant $\beta_1$, we will have attention weights represented by the following block matrix 
\begin{align*}
    A^{(1)} = \beta_1 \sqparen{\begin{matrix}
        0_d & 0_d & 0_d \\
        0_d& 0_d & 0_d \\
        0_d & I_d & 0_d
    \end{matrix}}
\end{align*}
where each $0_d$ is the $d \times d$ all-zeros matrix and $I_d \in \R^{d \times d}$ is the $d \times d$ identity matrix.
This implies pre-activations
\begin{align*}
    X {A}^{(1)} X^\top 
    &= \beta_1 \sqparen{\begin{matrix}
        0_d & 0_d \\
        I_d & 0_d
    \end{matrix} } \in \R^{T \times T}
\end{align*}
%
This implies activations
\begin{align*}
    \cS(X {A}^{(1)} X^\top) = \sqparen{\begin{matrix}
        1/(2d) \1 \1^\top & 1/(2d) \1_d \1_d^\top  \\
        I_d & 0_d \\
    \end{matrix}} 
\end{align*}
where $1/(2d) \1 \1^\top$ is the $d \times d$ constant matrix with entries $1/(2d)$.
%
Finally, we have layer-1 outputs
\begin{align*}
    \cS(X {A}^{(1)} X^\top) X =  \sqparen{
    \begin{matrix}
        1/(2d) \1 \1^\top(P + Y_P) & \dots & \dots  \\
        P & \dots & \dots 
    \end{matrix}}
\end{align*}
where we write $\dots$ for block matrices that will not factor into later computations and therefore do not matter.

The second attention layer computes $Y$.
Recall that the input to the second attention is 
\begin{align*}
    h^{(1)} &= 
    \sqparen{\begin{matrix}
        X & \cS(X {A}^{(1)} X^\top) X
    \end{matrix}}  \\
    &= 
    {\begin{bmatrix}
        P  & I_{d} & 0 &
        1/(2d) \1 \1^\top(P + Y_P) & \dots & \dots \\
        Y_P& 0 & I_{d} &
        P & \dots & \dots 
    \end{bmatrix}}
\end{align*}
Define a constant $\beta_2$.
We will have attention weights 
\begin{align*}
    {A}^{(2)} = \beta_2 \sqparen{\begin{array}{ccc|ccc}
        0 & 0 & 0 & 0 & 0 & 0  \\
        0 & 0 & 0 & I& 0 & 0 \\
        0 & 0 & 0 & 0& 0 & 0 \\
        \hline
        0 & 0 & 0 & 0& 0 & 0 \\
        0& 0  & 0 & 0& 0 & 0  \\
        0 & 0 & 0 & 0& 0 & 0 \\
    \end{array}}
\end{align*}
This implies pre-activations
\begin{align*}
    h^{(1)} {A}^{(2)} h^{(1) \top} &= \beta_2 \sqparen{\begin{matrix}
        I_d \\
        0_d
    \end{matrix}} \begin{bmatrix}
        1/(2d) \1 \1^\top(P + Y_P)^\top & P^\top
    \end{bmatrix}\\
    &=\beta_2 \sqparen{\begin{matrix}
        1/(2d) \1 \1^\top(P + Y_P)^\top & P^\top \\
        0 & 0
    \end{matrix}} 
\end{align*}
Note that $1/(2d) \1 \1^\top(P + Y_P)^\top < 1$  so, as $\beta_2 \to \infty$, we have activations
\begin{align*}
    \cS(h^{(1)} {A}^{(2)} h^{(1) \top}) 
    &= \sqparen{\begin{matrix}
        0 & P^\top \\
        1/(2d) \1 \1^\top & 1/(2d) \1 \1^\top
    \end{matrix}} 
\end{align*}
This yields layer-2 outputs
\begin{align*}
    \cS(h^{(1)} {A}^{(2)} h^{(1) \top}) h^{(1)} =   {\begin{bmatrix}
        \dots & \dots & \dots & \dots & \dots & \dots \\
        P^\top Y_P & \dots & \dots & \dots & \dots & \dots 
    \end{bmatrix}}
\end{align*}
Finally, notice $P^{\top} Y_P = Y$ so that block recovers the desired canonical order.
\end{proof}

\subsection{Proof of Theorem 3}
\label{app:proof_thm3}

\copyinputs*

\begin{proof}[Proof of \Cref{prop:copy_inputs}]
    Observe, since we pad the beginning of the input with a $\BOS$ token to be all zeros, that
    \begin{align*}
        X &= \embed(s_{1:T}) \\
        &= \sqparen{\begin{matrix}
            0 & 0_{1 \times T} \\
            \begin{matrix}
                P \\
                Y_P \\
                S 
            \end{matrix} & I_{T} \\
        \end{matrix}} \\
        &= \sqparen{\begin{matrix}
                0_{1 \times d}& 0_{1 \times d} & 0_{1 \times d} & 0_{1 \times d} \\
                P  & I_{d} & 0_{d}& 0_{d}\\
                Y_P& 0_{d} & I_{d}& 0_{d} \\
                S & 0_{d}& 0_{d} & I_{d}\\
        \end{matrix}}\in \R^{(T+1) \times (d + T)}
    \end{align*}
    
    The first layer copies $P$ into the third block-row of the residual stream.
    For constant $\beta_1$, we will have attention weights represented by the following block matrix 
    \begin{align*}
         {A}^{(1)} = \beta_1 \sqparen{\begin{matrix}
            0 & 0 & 0 & 0\\
            0 & 0 & 0 & 0\\
            0 & I_d & 0 & 0\\
            0 & 0 & 0 & 0
        \end{matrix}}
    \end{align*}
    where each $0$ is the $d \times d$ all-zeros matrix.
    This implies pre-activations
    \begin{align*}
        X {A}^{(1)} X^\top 
        &= \beta_1 \sqparen{\begin{matrix}
            0 & 0_{1 \times d} & 0_{1 \times d} & 0_{1 \times d}\\
            0_{d \times 1} & 0_d & 0_d & 0_d\\
           0_{d \times 1} & I_d & 0_d & 0_d\\
           0_{d \times 1} & 0_d & 0_d & 0_d
        \end{matrix} } \in \R^{(T + 1) \times (T + 1)}
    \end{align*}
    Let $\Delta^{(1)}$ be the lower triangular matrix with $\Delta_{ij} = 1/i$ for $j\leq i$.
    Let $\Delta^{(1)}$ be the $d+1 \times d+1$ matrix with $\Delta_{ij} = 1 / (2d+i+1)$.
    And as $\beta_1 \to \infty$,
    \begin{align*}
        \cS (\MASK (X{A}^{(1)} X^\top)) &= 
        \begin{bmatrix}
            \Delta^{(1)} && 0 & 0\\
            0_{d \times 1} & I_d & 0 & 0\\
            0_{d \times 1} & \Delta^{(2)} & \cdots & \cdots
        \end{bmatrix} 
    \end{align*}
    We will write dots when the values in a block do not matter for subsequent computations.
    This yields layer 1 outputs:
    \begin{align*}
        \cS (\MASK (X {A}^{(1)} X^\top)) X 
        &= {\begin{bmatrix}
            \Delta^{(1)} [0_{1 \times d}; P] &0_{1 \times d} & 0_{1 \times d} & 0_{1 \times d} \\
             \cdots & \cdots  & \cdots & \cdots \\
            P & 0_{d \times d} & 0_{d \times d} & 0_{d \times d} \\
            0_{d} & \cdots & \cdots & \cdots
        \end{bmatrix}}\in \R^{T \times (d + T)}
    \end{align*}
    
    The second layer will produce $P^\top$ in the activations so that $Y$ is written to the residual stream.
    Recall:
    \begin{align*}
        h^{(1)} &= \begin{bmatrix}
                0_{1 \times d} & 0_{1 \times d}& 0_{1 \times d} & 0_{1 \times d}&  \Delta^{(1)} [0_{1 \times d}; P] &0_{1 \times d} & 0_{1 \times d} & 0_{1 \times d} \\
                P  & I_{d} & 0_{d}& 0_{d} & \cdots &\cdots  & \cdots & \cdots \\
                Y_P& 0_{d} & I_{d}& 0_{d} & P & 0_{d \times d} & 0_{d \times d} & 0_{d \times d} \\
                0_{d} & 0_{d}& 0_{d} & I_{d}& 0_{d} & \cdots & \cdots & \cdots
        \end{bmatrix}
    \end{align*}
    For constant $\beta_2$, we will have attention weights represented by the following block matrix 
    \begin{align*}
        {A}^{(2)} = \beta_2 \sqparen{\begin{array}{cccc|cccc}
            0 & 0 & 0 & 0 & 0 & 0 & 0 & 0  \\
            0 & 0 & 0 & 0& 0 & 0 & 0 & 0 \\
            0 & 0 & 0 & 0& 0 & 0 & 0 & 0 \\
            0 & 0 & 0 & 0& I_d & 0 & 0 & 0 \\
            \hline
            0& 0  & 0 & 0& 0 & 0 & 0 & 0  \\
            0 & 0 & 0 & 0& 0 & 0 & 0 & 0 \\
            0 & 0 & 0 & 0 & 0 & 0& 0 & 0 \\
            0 & 0 & 0 & 0 & 0 & 0& 0 & 0 
        \end{array}}
    \end{align*}
    This implies pre-activations
    \begin{align*}
        h^{(1)} A^{(2)} h^{(1)\top} &= \beta_2 \begin{bmatrix}
        0_{1 \times d} \\
        0_d \\
        0_d \\
        I_d
        \end{bmatrix} \begin{bmatrix}
            0_{d \times 1} & \Delta^{(1)} [0_{1 \times d}; P] & P & 0_d
        \end{bmatrix}\\
        &= \beta_2 \begin{bmatrix}
            0 & 0_{1 \times d} & 0_{1 \times d} & 0_{1 \times d} \\
            0_{d \times 1} & 0_d & 0_d & 0_d \\
           0_{d \times 1} & 0_d & 0_d & 0_d \\
            0_{d \times 1} &\Delta^{(1)} [0_{1 \times d}; P] & P^\top & 0_d 
        \end{bmatrix}.
    \end{align*}
    Now, notice that all entries of $\Delta^{(1)} [0_{1 \times d}; P]$ are less than 1, since all non-zero entries of $P$ are 1 and all entries below the first row of $\Delta^{(1)}$ are less than 1.
    Thus, as $\beta_2 \to \infty$, we have activations
    \begin{align*}
        \cS ( \MASK (h^{(1)} A^{(2)} h^{(1)\top})) &= \begin{bmatrix}
            \dots & 0 & 0 \\
            \dots & \dots & 0 \\
            0 & P^\top & 0 
        \end{bmatrix}.
    \end{align*}
    This implies layer-2 outputs:
    \begin{align*}
        \cS ( \MASK (h^{(1)} A^{(2)} h^{(1)\top}))  h^{(1)} &= \begin{bmatrix}
            0_{1 \times d} & 0_{1 \times d} &0_{1 \times d} &0_{1 \times d} &0_{1 \times d} &0_{1 \times d} &0_{1 \times d} &0_{1 \times d} \\
            \dots & \dots & \dots & \dots & \dots & \dots & \dots & \dots \\
            \dots & \dots & \dots & \dots & \dots & \dots & \dots & \dots \\
            Y & \dots & \dots & \dots & \dots & \dots & \dots & \dots \\
        \end{bmatrix}
    \end{align*}
    which recovers $Y$ as desired.
\end{proof}

\end{document}